\pdfoutput=1
\documentclass[11pt]{article}

\usepackage{naacl2021}
\usepackage{times}
\usepackage{latexsym}

\usepackage{url}
\usepackage{subcaption}
\usepackage{comment}
\usepackage[pdftex]{graphicx}
\usepackage{latexsym}
\usepackage{booktabs}

\usepackage{amsmath} % aligned 1s
\usepackage{amssymb} % arrows
\usepackage{bm}% bold math 

\usepackage[T1]{fontenc}
\usepackage[utf8]{inputenc}

\usepackage{microtype}

\title{Context-aware Decoder for Neural Machine Translation\\
using a Target-side  Document-Level Language Model}

\author{Amane Sugiyama \\
  The University of Tokyo\thanks{\,\,\,Currently at Mitsubishi UFJ Morgan Stanley Securities} \\
  \texttt{sugi@tkl.iis.u-tokyo.ac.jp} \\\And
  Naoki Yoshinaga \\
  Institute of Industrial Science, \\
  The University of Tokyo \\
  \texttt{ynaga@iis.u-tokyo.ac.jp} \\}

\begin{document}
\maketitle
\begin{abstract}
Although many end-to-end context-aware neural machine translation models have been proposed to incorporate inter-sentential contexts in translation, these models  can be trained only in domains where
% in an end-to-end manner 
parallel documents with sentential alignments
% Because such document-level parallel data 
exist. 
% for training.
% document-level parallel data,
% we cannot perform an effective context-aware translation in most cases.
We therefore present a simple method to perform context-aware decoding with any pre-trained sentence-level translation model by using a document-level language model.
Our context-aware decoder is built upon sentence-level parallel data
% sentence-level parallel corpora 
and target-side document-level monolingual data.
% thus no document-level parallel data is needed.
    From a theoretical viewpoint, our core contribution is the novel representation of contextual information using point-wise mutual information between context and the current sentence.
We demonstrate the effectiveness of our method on English to Russian translation, by evaluating with \textsc{bleu} and contrastive tests for context-aware translation.
\end{abstract}

\section{Introduction}
Neural machine translation (\textsc{nmt}) has typically been explored in sentence-level translation settings.
Such sentence-level \textsc{nmt} models inevitably suffer from ambiguities when a source sentence has multiple plausible interpretations. Examples of such ambiguities include anaphora, ellipsis, and lexical coherence~\cite{voita-etal-2019-good}; although resolving these ambiguities has only a minor impact on the translation performance measured by \textsc{bleu} scores~\cite{papineni-etal-2002-bleu}, they are vital in smoothly reading the translated documents.

To address this issue, context-aware \textsc{nmt} models which incorporate document-level information in translation have recently been explored~\cite{jean2017does,wang-etal-2017-exploiting-cross,tiedemann-scherrer-2017-neural,maruf-haffari-2018-document,voita-etal-2018-context,bawden-etal-2018-evaluating,miculicich-etal-2018-document,maruf-etal-2019-selective,voita-etal-2019-good,yu-etal-2020-better,kang-etal-2020-dynamic,zhang-etal-2020-long}.
% %to address the issue in various research
\begin{comment}
Most of these models are end-to-end models that take as input the current source sentence to be translated and the context sentences, and then output a translation.
These models are trained on document-level parallel data, namely, sentence pairs with surrounding, usually preceding, sentences in the source and target language.
However, in practical scenarios, large-scale document-level parallel data is not available in most language pairs and domains,
%~\cite{sugiyama-yoshinaga-2019-data}
posing a challenge to building context-aware \textsc{nmt} systems~\cite{sugiyama-yoshinaga-2019-data}
 (\S~\ref{sec:related_work}).
\end{comment}
Most of these models are end-to-end models that require document-level parallel data with sentential alignments for training. However, this data is available in only a few 
% language pairs and 
domains~\cite{sugiyama-yoshinaga-2019-data}. Researchers have therefore started
to utilize target-side monolingual data to construct auxiliary models which help a sentence-level \textsc{nmt} model perform context-aware translation~\cite{voita-etal-2019-context,stahlberg-etal-2019-cued,yu-etal-2020-better}.

In this study, we propose a simple yet effective approach to context-aware \textsc{nmt} 
% consisting of 
using two primitive components, a sentence-level \textsc{nmt} model and a document-level language model (\textsc{lm}).
We can independently train the two components on common sentence-level parallel data and document-level monolingual data, respectively, without using document-level parallel data. Our approach thereby makes it possible to perform context-aware translation with any pre-trained sentence-level \textsc{nmt} model, using a pre-trained document-level \textsc{lm}.

% and thereby no document-level bilingual data is needed.
To give a probabilistic foundation to this combination of two independent models, we exploit
% take advantage of 
the probabilistic nature of \textsc{nmt} decoding.
When generating a sequence, a left-to-right decoder outputs a categorical probability distribution over the vocabulary at every time step.
% $t$.
The decoder assigns higher probabilities to the tokens that would be more suitable at that step.
Therefore, 
% we can assume that 
when multiple valid translations are possible for the source sentence,
% , which has ambiguities a sentence-level \textsc{nmt} is confused by, 
the decoder just gives a higher 
% sequence 
probability to the translation that is plausible without considering contexts. 
% than to wrong ones.
We thus adjust the probability distributions in a context-aware manner using a target-side document-level \textsc{\textsc{lm}} which 
% is capable of modeling 
models inter-sentential dependencies in the target-side document.

% Since a network structure of \textsc{nmt} models evolves very quickly, model-agnostic approach like ours is more preferable than model-tweaking approach (e.g. multi-encoder models).

We evaluate our methods on English to Russian translations with the OpenSubtitles2018 corpus~\cite{lison-etal-2018-opensubtitles2018} in terms of the \textsc{bleu} scores and contrastive discourse test sets~\cite{voita-etal-2019-good}. Experimental results confirm that our method achieved comparable performance with existing context-aware \textsc{nmt} models that require either document-level parallel data~\cite{zhang-etal-2018-improving,sugiyama-yoshinaga-2019-data} or more than one additional model~\cite{voita-etal-2019-context,yu-etal-2020-better} for capturing contexts in translation.

The contributions of this paper are as follows:
\begin{itemize}
    \item We theoretically derived \textsc{c-score}, a score to \textbf{qualify context-aware translation without the need for document-level parallel data}.
    \item Two formulations with \textsc{c-score} \textbf{turn any pre-trained sentence-level \textsc{nmt} model into a context-aware model}, if it generates $n$-best outputs or performs left-to-right decoding.
    %  in various language pairs and domains.
    \item A comparison between our approach and shallow fusion~\cite{gulcehre2015using} reveals that our approach \textbf{reformulates shallow fusion while adding a probabilistic foundation}.
%    \item our approach is based on \textsc{nmt} decoding combined with a language model, and therefore could be regarded as an extension of shallow fusion~\cite{garcia2017using}, but our approach is more sophisticated in that it has a probabilistic grounding;
%    \item through experiments, we confirmed that the proposed approach based on \textsc{pmi} computed by a language model highly contribute to improve performance on the contrastive tests, which are designed to evaluate the performance of context-aware \textsc{nmt} models.
\end{itemize}

\section{Context-aware Decoding using Document-level Language Model}

In this section, assuming a sentence-level encoder-decoder model~\cite{bahdanau2015,vaswani2017attention},
we first derive \textit{context-aware score} (\textsc{c-score} for short), a context-aware objective function of outputs to be maximized in decoding. We then describe how to compute the \textsc{c-score} using the decoder with a document-level language model (\textsc{d-lm}) (\S~\ref{sec:objective}). We finally detail how to perform context-aware decoding based on \textsc{c-score} (\S~\ref{sec:search}).

\subsection{\textsc{c-score}: objective function for context-aware \textsc{nmt} decoding}\label{sec:objective}
% \subsubsection{Derivation} % 
Let us consider the problem of finding a translation $\bm{y}$ of a source sentence $\bm{x}$ in a document.
The target-side context sentence(s)
% (possibly composed of multiple sentences) 
preceding $\bm{y}$, $\bm{c}^{(\bm{y})}$,
are to be given by the past translations.
% we hereafter refer to this target-side context $\bm{c}^{(\bm{y})}$ in translating $x$ as $c$, since we do not explicitly consider the source-side context.
We formulate context-aware translation conditioned on 
% $\bm{c}^{(\bm{y})}$ 
$\bm{c}^{(\bm{y})}$ as the maximization of the conditional probability $p(\bm{y}|\bm{x},\bm{c}^{(\bm{y})})$,
\begin{align}
    \hat{\bm{y}} &= \arg\max_{\bm{y}} \log p(\bm{y}|\bm{x},\bm{c}^{(\bm{y})}) \nonumber \\
    &= \arg\max_{\bm{y}} \log \frac{p(\bm{c}^{(\bm{y})}|\bm{x},\bm{y})p(\bm{y}|\bm{x})}{p(\bm{c}^{(\bm{y})}|\bm{x})} \nonumber \\
    &= \arg\max_{\bm{y}} \log p(\bm{c}^{(\bm{y})}|\bm{x},\bm{y})p(\bm{y}|\bm{x}). \label{eq:pyxc}
\end{align}
Assuming that $\bm{x}$ and $\bm{y}$ are semantically similar, we make the following approximation,
\begin{align}
    p(\bm{c}^{(\bm{y})}|\bm{y},\bm{x})\approx p(\bm{c}^{(\bm{y})}|\bm{y}). \label{eq:approx}
\end{align}
From Eq.~\ref{eq:pyxc} and Eq.~\ref{eq:approx}, we obtain
\begin{align}
    \hat{\bm{y}} &\approx \arg\max_{\bm{y}} \log p(\bm{c}^{(\bm{y})}|\bm{y})p(\bm{y}|\bm{x}) \nonumber \\
    &= \arg\max_{y} \log \frac{p(\bm{c}^{(\bm{y})}, \bm{y})}{p(\bm{c}^{(\bm{y})})p(\bm{y})} p(\bm{y}|\bm{x}) \nonumber \\
    &= \arg\max_{\bm{y}} \textsc{c-score}(\bm{y};\bm{x},\bm{c}^{(\bm{y})})\nonumber
\end{align}
where
\begin{align}
    \textsc{c-score}(\bm{y};\bm{x},\bm{c}^{(\bm{y})})= \log p(\bm{y}|\bm{x}) + \textsc{pmi}(\bm{c}^{(\bm{y})},\bm{y}) \label{eq:obj}\\
    \textsc{pmi}(\bm{c}^{(\bm{y})},\bm{y}) = \log \frac{p(\bm{c}^{(\bm{y})},\bm{y})}{p(\bm{c}^{(\bm{y})}) p(\bm{y})}
   = \log \frac{p(\bm{y}|\bm{c}^{(\bm{y})})}{p(\bm{y})} \label{eq:c_pmi}
\end{align}
$\textsc{pmi}(\bm{c}^{(\bm{y})},\bm{y})$ is the point-wise mutual information of $\bm{c}^{(\bm{y})}$ and $\bm{y}$ which represents the degree of co-occurrence of $\bm{y}$ and $\bm{c}^{(\bm{y})}$. % We hereafter consider \textsc{pmi} between a translation of a source sentence and the preceding translations (contexts).
% \subsubsection{Computation of \textsc{c-score}}\label{sec:comp_cs}
Given $\bm{x}$, $\bm{y}$ and $\bm{c}^{(\bm{y})}$, we can evaluate the \textsc{c-score} by computing the two terms in Eq.~\ref{eq:obj} using a sentence-level \textsc{nmt} (\textsc{s-nmt}) and a document-level \textsc{lm} (\textsc{d-lm}), respectively.

% \paragraph{Preliminary: neural sequence generation}
\paragraph{Notations} We first introduce some notation  
% that we use 
to explain the computation in Eq.~\ref{eq:obj} and Eq.~\ref{eq:c_pmi} using (auto-regressive) neural sequence generation models in \textsc{nmt} and \textsc{lm}.
% neural sequence generation models that are used in
% which include
% standard \textsc{nmt}s and \textsc{lm}s.
For a sequence $\bm{s}$ ($|\bm{s}|\ge 0$) and token $w$, a neural sequence generation model parameterized by $\theta$ can compute the log probability that $w$ follows $\bm{s}$, which we denote by $\log p_{\theta}(w|\bm{s}))$: % i.e
\begin{align}
    \log p_\theta(w\ \mathrm{folows}\ \bm{s}) = \log \frac{p_\theta(\bm{s}\cdot w)}{p_\theta(\bm{s})} =\log p_{\theta}(w|\bm{s}) \nonumber
\end{align}
where ``$\cdot$'' denotes sequence concatenation.
Applying this auto-regressively, for any sequence $\bm{s}^{(1)}$ ($|\bm{s}^{(1)}|\ge 0$) and $\bm{s}^{(2)}$ ($|\bm{s}^{(2)}|\ge 1$), the probability that $\bm{s}^{(2)}$ follows $\bm{s}^{(1)}$ is thereby computed as:
\begin{align}
    &\log p_\theta(\bm{s}^{(2)}\ \mathrm{follows}\ \bm{s}^{(1)}) \nonumber \\
    &\ = \log p_{\theta}(\bm{s}^{(2)}|\bm{s}^{(1)})
    =\sum_{t=1}^{|\bm{s}^{(2)}|} \log p_{\theta}(\bm{s}^{(2)}_t|\bm{s}^{(1)}\cdot \bm{s}^{(2)}_{<t}), \nonumber \\
    & \textrm{where}\ \bm{s}^{(2)}_{<t}=[s_1,\dots,s_{t-1}]. \label{eq:factorization}
\end{align}

\paragraph{$p(\bm{y}|\bm{x})$ computed by sentence-level \textsc{nmt}}
Computing $\log p(\bm{y}|\bm{x})$ using an \textsc{s-nmt} is straightforward.
Suppose $\bm{y}$ to be a sequence of raw tokens, $\bm{y}=[y_1,\dots,y_T]$.
Then $\log p(\bm{y}|\bm{x})$ is computed by
\begin{align}
    \log p(\bm{y}|\bm{x}) = \log p_{\textsc{s-nmt}}(\tilde{\bm{y}}; \bm{x}) \label{eq:tm_logp}
\end{align}
where $\tilde{\bm{y}} = [y_1,\dots,y_T, \texttt{</s>}]$ and $\texttt{</s>}$ is a special token to indicate the end of sentence.
% </s>を分離して考えるバージョン
%Suppose $y$ is a sequence of tokens $[y_1,\dots,y_T]$ without the start/end tokens (denoted by \texttt{<s>} and \texttt{</s>}, respectively).\bm{
%$\log p(\bm{y}|\bm{x})$ is factorized into token-wise conditional probabilities computed by the \textsc{nmt}.
%\begin{align}
%    \log p(\bm{y}|\bm{x}) &= \log p_{\textsc{s-nmt}}(y\cdot\texttt{</s>}|\texttt{<s>}; x) \nonumber \\
%    &= \sum_{t=1}^{T+1} \log p_{\textsc{s-nmt}}(\tilde{y}_t|\texttt{<s>}\cdot\tilde{y}_{<t}; x) \label{eq:tm_logp}
%\end{align}
%where $\tilde{y} = [y_1, \dots, y_T, \texttt{</s>}]$, $T$ is the input length, and the operator ``$\cdot$'' denotes sequence concatenation.

\paragraph{\textsc{pmi} computed by document-level \textsc{lm}}
To compute the components of $\textsc{pmi}(\bm{c}^{(\bm{y})},\bm{y})$, $p(\bm{y})$ and $p(\bm{y}|\bm{c}^{(\bm{y})})$, we use a document-level language model (\textsc{d-lm}) which can handle long text spans containing multiple sentences.

% During training, we 
We generate training examples for \textsc{d-lm} from a document as follows. We assume \textsc{d-lm} explicitly models sentence boundaries.
We first insert the special token \texttt{</s>} into every sentence boundary including the start and end of the document.
With this preprocessing, all the sentences start immediately after an \texttt{</s>} token and end immediately before an \texttt{</s>} token.
% ToDo: 確認。ここまで詳しく書かなくてよい？（査読ではspan同市がoverlapすることがあるかと聞かれた）
We then sample text spans from the document using a sliding window, where the start and end of the span do not have to match sentence boundaries.
The sliding window's size is larger than the stride size, so adjacent spans may overlap.
The resulting sequence is fed to the \textsc{d-lm} for training.
%We then randomly choose a text span of length $W$ from the document, where the start and end of the span do not have to match sentence boundaries, and the resulting sequence is fed to the \textsc{d-lm} for training.
Note that \texttt{</s>} for \textsc{d-lm} indicates sentence boundaries, in other words, both the start and end of the sequence.

Using \textsc{d-lm}, $p(\bm{y})$ is computed by
\begin{align}
    p(\bm{y}) = p_{\textsc{d-lm}}(\tilde{\bm{y}}|\texttt{</s>}). \label{eq:nlm}
\end{align}
where $\tilde{\bm{y}} = [y_1,\dots,y_T,\texttt{</s>}]$.

To compute $p(\bm{y}|\bm{c}^{(\bm{y})})$, we first obtain the context sequence $\tilde{\bm{c}}^{(\bm{y})}$ by concatenating all the sentences in $\bm{c}^{(\bm{y})}$ with \texttt{</s>}. 
% inserted at every sentence boundary.
We then compute the conditional probability $p(\bm{y}|\bm{c}^{(\bm{y})})$ by
\begin{align}
    p(\bm{y}|\bm{c}^{(\bm{y})}) = p_{\textsc{d-lm}}(\tilde{\bm{y}}|\tilde{\bm{c}}^{(\bm{y})}) \label{eq:clm}
\end{align}
where $\tilde{\bm{y}}=[y_1,\dots,y_T,\texttt{</s>}]$.

Let us explain why we use the boundary-aware \textsc{d-lm} rather than boundary-agnostic \textsc{d-lm}.\footnote{We cannot rely on punctuations
% such as ``.'' and ``?''
to know sentence boundaries, since they can be omitted in some domains.}

% whereas boundary-agnostic \textsc{d-lm},
% such as \textsc{bert}~\cite{devlin2018bert}, 
% which do not use a marker (\texttt{</s>}) to explicitly tell sentence boundaries, are common.
%  (\textit{e.g.}, subtitles)
%\footnote{\textsc{lm}s may be implicitly aware of sentence boundaries based on punctuations such as ``.'' and ``?''. However, 
% corpora in some languages and domains such as subtitles contain sentences with such punctuation omitted, 
%since punctuation can be omitted in some domains (\textit{e.g.}, subtitles), 
% in general 
%we cannot rely on it to let the \textsc{lm} know sentence boundaries.}
% We still can recognize the sentence boundaries in those corpora by, for example, timestamps of subtitles and therefore can split them into sentences during preprocessing.}
Firstly, boundary-agnostic \textsc{lm}s cannot compute the probability that a sentence is closed with a certain length, namely, Eq.~\ref{eq:nlm} cannot be computed.
Secondly, they also cannot compute $p(\bm{y}|\bm{c}^{(\bm{y})})$ correctly.
For example, suppose the context $\bm{c}^{(\bm{y})}$ is ``he's my friend'' (with the punctuation ``.'' omitted), and the current target sentence $\bm{y}$ is ``he's nice.''
In this case, Eq.~\ref{eq:clm} is computed by
\begin{align}
    p(\bm{y}|\bm{c}^{(\bm{y})}) =p_{\textsc{d-lm}}([\textrm{he,'s,nice}] | [\textrm{he,'s,my,friend}]). \nonumber
\end{align}
However, this estimation of $p(\bm{y}|\bm{c}^{(\bm{y})})$ can underestimate
% may be much less than 
the actual $p(\bm{y}|\bm{c}^{(\bm{y})})$ because Eq.~\ref{eq:clm} inevitably gives significant probabilities to other $\bm{y}$ such as ``'s father'' as well, since ``He's my friend's father'' is fluent as a sequence.
This behavior is unsuitable for $\bm{y}$,\footnote{Strictly speaking, we assume $\bm{y}$ to be a realization of a random variable $Y$ which is a sentence sampled from the space of an infinitely large document.} since ``'s father'' is not a complete sentence.

\subsection{Searching for the optimal solution}\label{sec:search}
Searching for the optimal output $\bm{y}$ that maximizes the \textsc{c-score} 
% presented in the previous section 
is not trivial since there are $\mathcal{O}(V^T)$ candidate sequences where $V$ is the vocabulary size and $T$ is the maximum length of sequences to be searched.
We investigate two approaches to obtain approximate solutions: reranking (\S~\ref{sec:reranking}) and context-aware beam search (\S~\ref{sec:bs}).

\subsubsection{Reranking with \textsc{c-score}}\label{sec:reranking}
We first generate $B$ hypotheses of the translation $\mathcal{H}_B = \{\bm{y}^1, \dots, \bm{y}^B\}$ with beam search of beam size $B$ using the sentence-level \textsc{nmt} model.
We then choose the one that maximizes the \textsc{c-score}.
\begin{align}
    \hat{y} = \arg\max_{\bm{y}\in \mathcal{H}_B} \textsc{c-score}(\bm{y};\bm{x},\bm{c}^{(\bm{y})})
\end{align}

An issue with reranking is that we need to set $B$ to a large value when the diversity of models' outputs is limited~\cite{yu-etal-2020-better}, which increases the cost of decoding. We therefore attempt to integrate \textsc{c-score} into the decoding with beam search.

%A potential issue of this approach is low diversity in the hypotheses.
%It is known that hypotheses obtained by beam search tend to share tokens in their front part, namely, the variation 
%% in those sentences 
%is unevenly concentrated on the tail.
%Moreover, 
%% semantically important 
%plausible variations can be pushed out of the hypotheses by trivial variations such as whether to put a comma.
%Therefore, ambiguity to be disambiguated using contextual information may be pruned in beam search. 
% We should mention that reranking is yet useful when we want to perform context-aware decoding with recent non-autoregressive \textsc{nmt}~\cite{gu2014nonauto}.

\subsubsection{Context-aware beam search}\label{sec:bs}
Context-aware beam search (\textsc{c-aware} beam) is beam search that is extended to work with \textsc{c-score}.
\textsc{c-score} (Eq.~\ref{eq:obj}) can be decomposed into token-wise \textsc{c-score}s (Eq.~\ref{eq:factorization} through Eq.~\ref{eq:clm}).
\begin{align}
    \textsc{c-score}(\bm{y};\bm{x},\bm{c}^{(\bm{y})}) \nonumber 
    &= \log p(\bm{y}|\bm{x}) + \textsc{pmi}(\bm{c}^{(\bm{y})},\bm{y}) \nonumber \\
    &= \sum_{t=1}^{T+1} \textsc{c-score}_{w}(\tilde{y}_t|\tilde{\bm{y}}_{<t})
\end{align}
where
\begin{align}
    \textsc{c-score}_{w}(\tilde{y}_t|\tilde{\bm{y}}_{<t})      = &
        \log p_{\textsc{s-nmt}}(\tilde{y}_t|\tilde{\bm{y}}_{<t}; x) \nonumber \\
        & + \log \frac{p_{\textsc{d-lm}}(\tilde{y}_t|\tilde{\bm{c}}^{(\bm{y})}\cdot\tilde{\bm{y}}_{<t})}{p_{\textsc{d-lm}}(\tilde{y}_t|\texttt{</s>}\cdot\tilde{\bm{y}}_{<t})} & \label{eq:tok_cs}
\end{align}
By this decomposition, 
% the token-wise 
\textsc{c-score}$_w$ 
% at time step $t$
is conditioned on the partial sequence generated by time step $t$. We can therefore apply beam search to generate sequences in an auto-regressive manner.

The first term of 
% the token-wise \textsc{c-score}
Eq.~\ref{eq:tok_cs} represents the translation probability for the $t$-th token.
The second term can be interpreted as \textsc{pmi} between the $t$-th token and the context, that is, how consistent the $t$-th token is with the context.
Compared to the reranking approach, \textsc{c-aware} beam can be considered to maximize the \textsc{c-score} more directly in the sense that disambiguation and token selection based on the context are performed at every step in beam search.
Thus \textsc{c-aware} beam will more space-efficiently consider diverse hypotheses with the same beam size $B$ than \textsc{c-aware} rerank.
% alleviate the low diversity problem of reranking when using the same beam size $B$.

\subsubsection{Smoothing probabilities for \textsc{pmi}} \label{sec:smoothing}
In our preliminary experiments, we observe that the original \textsc{c-aware} beam significantly improves contrastive tests but deteriorates \textsc{bleu} at the same time.
By analyzing contextual \textsc{pmi} correlation between source and target texts, we find the \textsc{pmi} term in the \textsc{c-score} sometimes takes an excessively large value against the translation probability term, which destroys the \textsc{c-score}.
This is understood intuitively by the fact that the calculation of \textsc{pmi} includes subtraction of log probability, and log probability may take a very small negative value to represent a probability close to zero.

To alleviate this problem, we adopt a smoothing method for probabilities.
% including probability smoothing and restriction of the vocabulary set on which \textsc{lm} gives probabilities.
For simplicity, in this paper, we only present the temperature scaling ($T$-scaling, for short)~\cite{pmlr-v70-guo17a}.
%We explore two smoothing methods which we found simple but effective.
%These smoothing techniques are applied to both the % conditional (
%numerator and  denominator
$T$-scaling replaces $p_{y=w}$ by
\begin{align}
    \bar{p}_{y=w} = \frac{p_{y=w}^{1/T}}{\sum_{w'}p_{y=w'}^{1/T}} 
\end{align}
where $T$ is a hyper-parameter.
$T=1$ is equivalent to no smoothing. We choose $T$ from $[1,\infty)$ to flatten the probability distribution.
$T$-scaling is applied to both the numerator and denominator using the same $T$.

%\smallskip\noindent\textbf{Add-$\alpha$ smoothing} is inspired by the count smoothing with the same name, and replaces $p_{y=w}$  by
%\begin{align}
%    \bar{p}_{y=w} = \frac{p_{y=w} + \frac{\alpha}{|V|}}{\alpha + 1} 
%\end{align}
%where $\alpha$ is a hyper-parameter to be tuned to maximize the translation performance on a development set.
%Smaller probabilities take larger scaling ($\bar{p}_{y=w}/p_{y=w}$).
%$\alpha=0$ is equivalent to no smoothing and $\alpha=\infty$ is equivalent to setting all the token probabilities to $\frac{1}{|V|}$.

\subsubsection{On the relation to  shallow fusion} \label{sec:shallow_fusion}
Shallow fusion~\cite{gulcehre2015using} is a method to integrate probability distribution outputs obtained by \textsc{nmt} and \textsc{lm} at sentence level to form a new translation objective that is expected to promote fluency of translations.
The original shallow fusion score is computed using a sentence-level \textsc{nmt} (\textsc{s-nmt}) and language model (\textsc{s-lm}).
The token-wise formula of the computation is
\begin{align}
    \log p(y_t) &=\log p_{\textsc{s-nmt}}(y_t;\bm{x}) + \beta \log p_{\textsc{s-lm}}(y_t), 
\end{align}
where $\beta$ is a hyper-parameter.
In our notation with the document-level \textsc{lm}, this is written as
\begin{align}
    \log p(y_t) &= \log p_{\textsc{s-nmt}}(\tilde{y}_t|\tilde{\bm{y}}_{<t}; \bm{x}) \nonumber \\
    & \quad + \beta\log p_{\textsc{s-lm}}(\tilde{y}_t|\texttt{</s>} \cdot\tilde{\bm{y}}_{<t}). 
\end{align}
A natural extension of this objective to the context-aware scenario should be
\begin{align}
    p(y_t|\bm{c}^{(\bm{y})}) &= \log p_{\textsc{nmt}}(\tilde{y}_t|\tilde{\bm{y}}_{<t}; x) \nonumber \\
    &\quad + \beta\log p_{\textsc{d-lm}}(\tilde{y}_t|\tilde{\bm{c}}^{(\bm{y})}\cdot\tilde{\bm{y}}_{<t}),
\end{align}
where context $\tilde{c}^{(\bm{y})}$ is integrated into the condition.
We call this \textit{conditional (document-level) shallow fusion}.
Obviously, this is what we obtain from Eq.~\ref{eq:tok_cs} by ignoring the discount of the unconditional \textsc{lm} probability $p_{\textsc{d-lm}}(\tilde{y}_t|\texttt{</s>}\cdot \tilde{y}_{<t})$.

% An important 
% potential 
% difference made by
Due to the absence of discounting with the unconditional \textsc{lm},
% the conditional document-level shallow fusion
conditional shallow fusion 
would
prefer tokens which frequently occur regardless of the context.
% In this sense the \textsc{c-score}-based approaches ...
It is also worth noting that, when the context is empty, conditional shallow fusion falls back to the original shallow fusion, whereas our \textsc{c-score} falls back to sentence-level \textsc{nmt}. Therefore, we view \textsc{c-score} as a reformulation of shallow fusion for context-aware translation.

\section{Experimental Setup}
We evaluate our methods on English to Russian
% (En-Ru)
translation, in terms of \textsc{bleu} scores~\cite{papineni-etal-2002-bleu} and contrastive tests~\cite{voita-etal-2019-good}.

\subsection{Datasets and preprocessing}
We use the 
% publicly available 
OpenSubtitles2018 corpus~\cite{lison-etal-2018-opensubtitles2018} for parallel  and monolingual data.
Following the criteria for document segmentation and filtering on sentence pairs presented by \cite{voita-etal-2019-good}, we build monolingual and parallel data as follows.
To build monolingual data, we add document boundary information into each document such that they consist of contiguous subtitle sentences from the same movie and the timestamp difference of any two adjacent sentences is no more than seven seconds. 
To build parallel data, we pick subtitle pairs where the time overlap between the source and target language subtitles is at least 0.9 (to reduce alignment errors).
% and use these 
% as sentence pairs.
% we pick sentence pairs with a relative time overlap of subtitle frames between source and target languages being at least $0.9$ to reduce alignment errors.
For the training of multi-encoder \textsc{nmt} models, document boundary information is added to the parallel data based on the source-side timestamps as with the monolingual data.
Prior to building the Russian data, we remove the movies from which the  contrastive test sets (\S~\ref{sec:metrics}) were made.

\begin{comment}
    \begin{table}[t]
        \small
                \centering
    
            \begin{tabular}{c|ccc} \toprule
                 & French & Russian & English \\ \midrule
                train & 106M, 8.61 & 44.8M, 8.00 & 440M, 8.24 \\
                dev & 41.7k, 8.16 & 33.2k, 8.27 & 49.8k, 8.49 \\
            \bottomrule
            \end{tabular}
            \caption{1 the monolingual data (\# of sentences, \# of tokens per sentence).}
            \label{tab:stats_mono_data}
      \end{table}
      
    \begin{table}[t]
    \small
            \centering
            \begin{tabular}{@{}c@{\,\,}|@{\,\,}c@{\,\,\,\,}c@{\,\,\,\,}c@{}} \toprule
                 & En-Fr & En-Ru & Ja-En \\ \midrule
                train & 17.6M, 10.5, 10.7 & 12.5M, 10.2, 9.6 & 742k, 8.92, 10.0 \\
                dev & 1665, 11.8, 12.3 & 4407, 12.3, 11.8 & 2834, 11.9, 11.8 \\
                test & 4166, 12.6, 12.7 & 4439, 11.7, 11.1 & 2820, 11.3, 11.4 \\
            \bottomrule
            \end{tabular}
            \caption{Statistics of the monolingual data (\# of sentences and \# of tokens per source and target sentence).}
            \label{tab:stats_para_data}
    \end{table}
    
\end{comment}

We perform punctuation normalization, tokenization, and truecasing on the source and target texts
% English and Russian text 
using Moses toolkit v4.0.\footnote{\url{http://www.statmt.org/moses/}}
We then encode the texts into subwords using SentencePiece (v0.1.81)\footnote{\url{https://github.com/google/sentencepiece}} with unigram \textsc{lm}.
 The subword vocabularies are of 16,000 tokens and 
% separately 
trained for each language.
The statistics of the datasets are listed in Table~\ref{tab:stats_mono_data}.

\subsection{Models}
We compare our methods to one sentence-level translation model (\textbf{SentTransformer})~\cite{vaswani2017attention} and three context-aware translation models: Document transformer~\cite{zhang-etal-2018-improving}, DocRepair~\cite{voita-etal-2019-context}, and Bayes Document Reranker~\cite{yu-etal-2020-better}. All the context-aware models use the previous three sentences as context. 

\smallskip\noindent\textbf{Document Transformer} (\textbf{DocTransformer}, for short) is a multi-encoder document-level \textsc{nmt} model which takes source-side context as an auxiliary input and can be thus trained from document-level parallel data.
We follow \citep{zhang-etal-2018-improving}'s configuration for DocTransformer. 

\begin{comment}
\begin{table}[t]
    \small
        \centering
        \begin{tabular}{c|cc}
            \toprule
             & Parallel & Monolingual \\
             \midrule
             Train & 5.8M. 9.9, 9.4 & 30M, 8.5 \\
             Dev & 6.0k, 10.1, 9.6 & 23k, 8.9 \\
             Test-1 &  4.7k, 10.3, 9.7  & - \\ 
             Test-2 &  4.7k, 9.3, 8.8 & - \\ 
             Test-3 &  4.3k, 9.9, 9.0 & - \\ 
             \bottomrule
        \end{tabular}
%        \begin{tabular}{c|ccc} \toprule
%            & Train & Dev & Test \\ \midrule
%            Mono. & 30M, 8.5 & 23k, 8.9 & - \\
%            Para. & 5.8M, 9.9, 9.4 & 6.0K, 10.1, 9.6 & 14k, 9.8, 9.1 \\
%        \bottomrule
%        \end{tabular}
        \caption{Statistics of the monolingual and parallel data (\# of sentences, \# of tokens per sentence (source and target for the parallel data)).}
        \label{tab:stats_mono_data}
  \end{table}
\end{comment}

\begin{table}[t]
    \small
        \centering
        \begin{tabular}{@{\,}l@{\,\,\,\,}c@{\,\,\,}c@{\,\,}c@{\,\,\,\,}c@{\,\,\,}c@{\,\,}c@{\,\,\,\,}c@{\,\,\,}c@{\,}}
            \toprule
             & 
             \multicolumn{3}{c@{\,\,\,\,}}{\textbf{Train}} & 
             \multicolumn{3}{c@{\,\,\,\,}}{\textbf{Dev.}} & 
             \multicolumn{2}{c}{\textbf{Test}} \\
             \cmidrule(r{.5em}){2-4}
             \cmidrule(l{.25em}r{.5em}){5-7}
             \cmidrule(r{.25em}){8-9}
%              &  \multicolumn{2}{@{\,\,}c@{\,\,\,\,}}{para.} & mono & \multicolumn{2}{@{\,\,}c@{\,\,\,\,}}{para.} & mono & \multicolumn{2}{@{\,\,}c@{\,}}{para.} \\
             & src & trg & (mono) & src & trg & (mono) & src & trg \\
             \midrule
             \# sentences & \multicolumn{2}{c}{5.8M} & 30M & \multicolumn{2}{c}{6.0k} & 23k & \multicolumn{2}{c}{15.5k} \\
             avg. \# tokens & 9.9 & 9.4 & 8.5 & 10.1 & 9.6 & 8.9 & 9.8 & 9.1 \\ 
             \bottomrule
        \end{tabular}
%        \begin{tabular}{c|ccc} \toprule
%            & Train & Dev & Test \\ \midrule
%            Mono. & 30M, 8.5 & 23k, 8.9 & - \\
%            Para. & 5.8M, 9.9, 9.4 & 6.0K, 10.1, 9.6 & 14k, 9.8, 9.1 \\
%        \bottomrule
%        \end{tabular}
        \caption{Statistics of the parallel and monolingual data.}
        % (\# of sentences, \# of tokens per sentence (source and target for the parallel data)).}
        \label{tab:stats_mono_data}
  \end{table}

\smallskip\noindent\textbf{DocRepair} is a sequence-to-sequence post-editing model.
It repairs document-level inconsistencies in a text, each sentence of which has been translated separately by a sentence-level \textsc{nmt} model.
DocRepair is trained on a pseudo parallel data made by pairing a monolingual corpus and its round-trip translations obtained using a back-translation model and a forward-translation model.

\smallskip\noindent\textbf{Bayes Document Reranker} (hereafter, \textbf{Bayes DocReranker}) performs document-level translation on a document containing $D$ sentences in the following steps.
First, it produces $B$-best translations for each sentence in the document and then produces a lattice of width $B$ and depth $D$, where each node corresponds to a candidate sentence.
It then performs document-level beam search of beam size $B'$ on the lattice using the following score: % function to obtain the optimal sequence of sentences.
\begin{align}
    \mathrm{S}&\mathrm{core}(\bm{y}_i; \bm{y}_{<i}, \bm{x}_i) 
        =  \nonumber \\
        & p_{\textsc{d-lm}}(\bm{y}_i | \bm{y}_{<i})   + \mathrm{Score}(\bm{y}_{i-1};\bm{y}_{<i-1}, \bm{x}_{i-1}) \nonumber \\
     & + \lambda_1 p_{\textsc{nmt}}(\bm{y}_i | \bm{x}_i) + \lambda_2 p_{\textsc{back-nmt}}(\bm{x}_i | \bm{y}_i) + \lambda_3 |\bm{y}_i|
\end{align}
Note that this document-level beam search is equivalent to the reranking procedure (\S~\ref{sec:reranking}) when $B'=1$.
% and, conversely, the reranking with \textsc{c-score} can be extended to the document-level beam search.
Therefore, the essential difference between Bayes DocReranker and our \textsc{c-score} reranking is the score function.

% The post-editing model of DocRepair and the sentence-level \textsc{nmt} models used for data augmentation and the first stage translation for our methods, DocRepair, and Bayes DocReranker are based on the same configuration of Transformer base (see~\cite{vaswani2017attention} for hyperparameter settings).
SentTransformer, the post-editing model of DocRepair, and the back-translation models are based on the same configuration of Transformer base (see~\cite{vaswani2017attention} for hyperparameter settings). The SentTransformer is trained using the 5.8M % English-Russian 
sentence pairs and is also used as the sentence-level \textsc{nmt} model in DocRepair, Bayes DocReranker, and our methods. For the training of DocTransformer, we use the 5.8M sentence pairs with document-level source context, which share the target-side sentences with the training data of
SentTransformer.
% the sentence-level \textsc{nmt} models.
Consequently, scores obtained from the model are for reference.\footnote{Although we can train DocTransformer only on pseudo document-level parallel data generated by back-translation, we confirmed in preliminary experiments that the resulting model exhibited poor performance.}
We also evaluate DocTransformer and SentTransformer using back-translation (\textsc{bt})~\cite{sennrich-etal-2016-improving} with the same monolingual data as the other models.

% for a fair comparison.
% \footnote{Bayes DocReranker and our proposed methods can use more than three sentences as context.}
We use no pre-existing  document-level parallel data to train the neural networks of DocRepair, Bayes DocReranker, and our methods, although we use a small amount of document-level parallel data as the development set to tune hyperparameters in the methods that combine multiple models.
Instead, document-level information is fed to the models via the round-trip augmented data (DocRepair) or language models (Bayes DocReranker and our methods).

\paragraph{Hyper-parameters}
We tune the models' hyper-parameters 
% $T$ (\S~\ref{sec:smoothing}) and $\beta$ (\S~\ref{sec:shallow_fusion})
based on \textsc{bleu} score on the development set in the evaluation with \textsc{bleu},
while we tune these hyper-parameters
%on the scores of contrastive tests
in the evaluation of contrastive tests
by maximizing the coefficient of \textsc{d-lm} under the constraint that it does not deteriorate \textsc{bleu} compared to the SentTransformer.
% baseline sentence-level decoder.

% We tune $T$ for each model (\textsc{c-aware} reranking and  beam), and tune $\beta$ for model (conditional and unconditional shallow fusion). $T$ is tuned in the range $[1, 16]$, and $\beta$ is tuned in the range $(0, 1)$.
% Since $\alpha$ and $T$ are both for smoothing, for simplicity, we do not apply them simultaneously to the models.
\begin{table}[t]
    \small
    \centering
    \begin{tabular}{@{\,\,}l@{\,\,\,\,}c@{\,\,\,\,\,\,}c@{}l@{\,\,\,\,}c@{}l@{\,\,}c@{}l@{\,}} \toprule
 \textbf{Models} & \textbf{para} & \multicolumn{5}{c}{\textbf{monolingual data}} \\
        & \textbf{only} & 6M && 15M && 30M &\\ \midrule
        SentTransformer (w/ \textsc{bt}) & 32.36 & 32.32 && 32.40 && 32.40& \\
         Shallow Fusion
            & n/a & 32.39 && 32.56 && 32.52& \\
        \midrule
        \multicolumn{4}{@{}l}{\textit{baselines}} \\
        DocTransformer  (w/ \textsc{bt}) & 32.50 & 32.36 && 31.88& & 31.59& \\
        DocRepair & n/a & 32.13 && 32.36 && 32.35 & \\
        Bayes DocReranker  & n/a & \textbf{32.80}&$^{*}$ & \textbf{33.58}&$^{**}$ & \textbf{33.75}&$^{**}$ \\
        \hfill w/o context & n/a & 32.53& & 33.44&$^{**}$ & 33.67&$^{**}$ \\
        \multicolumn{4}{@{}l}{\textit{proposed}} \\
        \textsc{c-aware} Rerank & n/a & 32.74&$^{*}$ & 33.01&$^{**}$ & 32.93&$^{*}$ \\
        \textsc{c-aware} Beam & n/a & 32.26& & 32.28& & 32.27& \\
        Cond. Shallow Fusion
            & n/a & 32.38 && 32.55 && 32.55& \\
         \bottomrule
    \end{tabular}
    \caption{Test set \textsc{bleu} scores. `*' and `**' indicate that gains from SentTransformer in the same column are statistically significant ($p<0.05$ and $p<0.01$) by bootstrap resampling with 1000 samples, respectively.}
    \label{tab:main_bleu}
\end{table}
For beam search to produce $B$-best outputs in Bayes DocReranker and our \textsc{c-aware} Rerank, we use a beam size of $B=20$.
For document-level beam search % on %the lattice 
of Bayes DocReranker, we use a beam size $B'=5$. For beam search of SentTransformer, DocTransformer, \textsc{c-aware} beam, and shallow fusion, we use a beam size of $B=4$.
%We confirmed on the dev set  that SentTransformer and DocTransformer models with beam size of $4$ perform better than or at least comparable to models with beam size of $10$. 
% We use three previous sentences as context for the context-aware models (2-to-2, \textsc{c-aware} beam, and conditional shallow fusion). 

\subsection{Document-level Language models}
The architecture of the document-level \textsc{lm} is the decoder part of a Transformer.
The number of decoder blocks is $12$.
The model size is $768$ with $12$ attention heads, and the inner layer of the feed-forward networks has $3072$ units.
We use position embeddings to represent position information.

As described in \S~\ref{sec:objective}, when training the language models, a special control symbol \texttt{</s>} is inserted at every sentence boundary.
Each training mini-batch contains text spans each of which is a randomly sampled fragment of a document with a maximum span length of $W=384$.
Text spans are batched such that about 32,000 tokens are in a training batch.

\subsection{Evaluation methods}\label{sec:metrics}
%We adopt contrastive tests~\citep{bawden-etal-2018-evaluating,voita-etal-2019-good,sugiyama-yoshinaga-2019-data}
The existing automatic metrics are not adequate to evaluate gains from additional contexts~\cite{bawden-etal-2018-evaluating,laubli-etal-2018-machine,muller-etal-2018-large,voita-etal-2019-good,sugiyama-yoshinaga-2019-data}. We thus adopt a contrastive test set~\citep{voita-etal-2019-good}
% for context-aware \textsc{nmt} 
to evaluate the model's ability to capture contextual information in translation, in addition to the evaluation by \textsc{bleu} scores~\cite{papineni-etal-2002-bleu} to confirm that the methods do not sacrifice general translation performance. \textsc{bleu} is computed using \texttt{multi-bleu.perl} from the Moses Toolkit after decoding the subword representation of the models' outputs into words using SentencePiece.

The contrastive test set consists of contrastive questions for context-aware \textsc{nmt} models to answer.
Each question has a source sentence $\bm{x}$, a source context $\bm{c}^{(\bm{x})}$, a target context $\bm{c}^{(\bm{y})}$, and translation candidates $\mathcal{Y}=\{y^1,\dots,y^M\}$.
Models must answer with a candidate $\hat{y}\in\mathcal{Y}$ which would be the most appropriate translation of $\bm{x}$, i.e.
\begin{align}
    \hat{y} = \arg\max_{y\in\mathcal{Y}} p(\bm{y}|\bm{x},\bm{c}^{(x)},\bm{c}^{(\bm{y})}) \nonumber
\end{align}
The test sets consist of 6000 examples in total.

% Translation quality is evaluated with tokenized \textsc{bleu}. 

\section{Results and Analysis}

\begin{table}[t]
    \small
    \centering
    \begin{tabular}{@{\,\,}l@{\,\,\,}r@{\quad}r@{\,\,\,}r@{\quad}r@{\,}} \toprule
        % Test name
        \textbf{Models}
            & \textbf{deixis}
            & \textbf{lex.c}
            & \textbf{ell.infl}
            & \textbf{ell.vp} \\
        \midrule
        SentTransformer
%           & $50$ & $45.86$ & $53.2$ & $27$ \\
            & 50.0 & 45.9 & 53.2 & 27.0 \\
            \hfill w/ \textsc{bt} & 50.0 & 45.9 & 51.6 & 26.8 \\
        \midrule
        \multicolumn{5}{@{}l}{\textit{baselines}} \\
        Doc-Transformer
            & 50.0 & 45.9 & 56.0 & 57.2 \\
        \hfill w/ \textsc{bt}
            & 50.0 & 45.9 & 64.4 & 68.2 \\
        DocRepair
%            & 89.08 & 75.8 & 82.2 & 67.2 \\
            & \textbf{89.1} & 75.8 & \textbf{82.2} & 67.2 \\
        Bayes DocReranker
%            & 65.24 & 72.2 & 59.6 & 44.6 \\
            & 65.2 & 72.2 & 59.6 & 44.6 \\
         \multicolumn{5}{@{}l}{\textit{proposed}} \\
        \textsc{c-score} & 86.9 & \textbf{94.9} & 78.2 & \textbf{77.0} 
        \\ %(T=4.0)
%            & 63.6 & 75.7 & 61.4 & 44.6 \\
%        \textsc{c-score} (T=2.0)
%            & 76.32 & 92.2 & 71.6 & 61.2 \\
%            & 76.3 & 92.2 & 71.6 & 61.2 \\
%        \textsc{c-score} (T=1.2)
%            & 84.8 & \mathbf{94.4} & 75.4 & \mathbf{74.0} \\
        %\hfill add-\alpha & 80 && 65&.5 & 81&.1 & 89&.2 & 75&.6 & 64&.6 & \mathbf{84} \\
        %\hfill T-scaling & 83 && \mathbf{69}&\mathbf{.5} & 85&.8 & \mathbf{94}&\mathbf{.2} & \mathbf{84}&\mathbf{.2} & 74&.4 & \mathbf{84} \\
        Cond. Shallow Fusion % $p(\bm{y}|\bm{c}^{(\bm{y})})$
        & 54.7 & 55.3 & 53.4 & 32.4 \\ 
        \midrule
        \textsc{d-lm} \hfill $\textsc{pmi}(\bm{c}^{(\bm{y})},y)$
%            & \mathbf{96.76}
            & \textbf{96.8}
            & \textbf{97.8}
            & 75.8
            & \textbf{90.6} \\
        \hfill $p(\bm{y}|\bm{c}^{(\bm{y})})$
%            & 89.72
            & 89.7
%            & 95.73
            & 95.7
            & \textbf{77.4}
            & 81.6 \\
        \bottomrule
    \end{tabular}
    \caption{Results on contrastive test sets.}
    \label{tab:disco}
\end{table}

\begin{figure*}[t]
    \begin{subfigure}{0.23\textwidth}
        \centering
        \includegraphics[width=\linewidth]{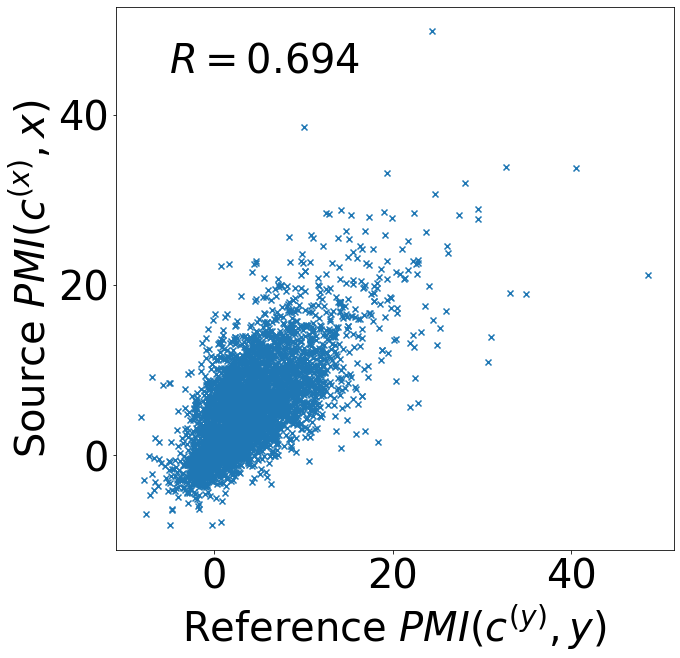}
        \caption{\textsc{pmi}}
        \label{fig:corr_pmi}
    \end{subfigure}
    \begin{subfigure}{0.24\textwidth}
        \centering
        \includegraphics[width=\linewidth]{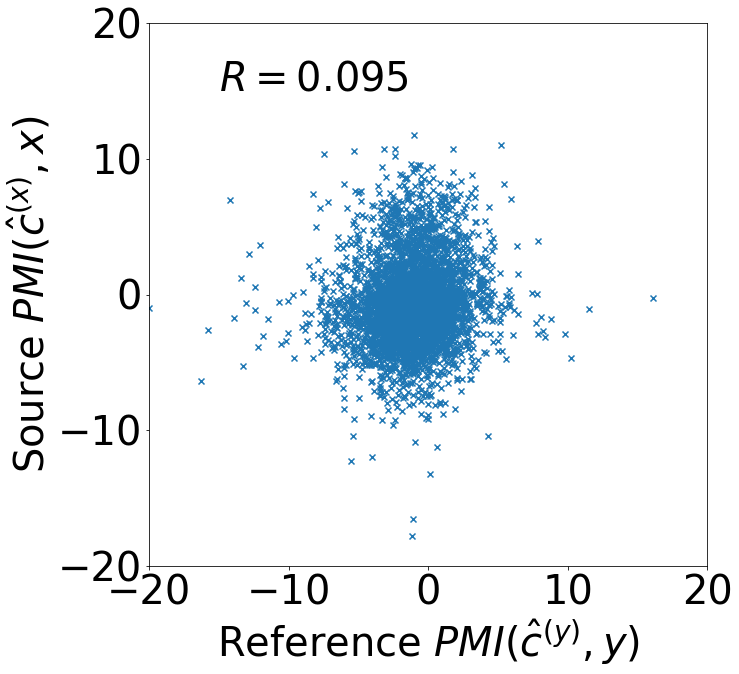}
        \caption{\textsc{pmi} with rand.~context}
        \label{fig:corr_pmi_false}
    \end{subfigure}
    \begin{subfigure}{0.24\textwidth}
        \includegraphics[width=\linewidth]{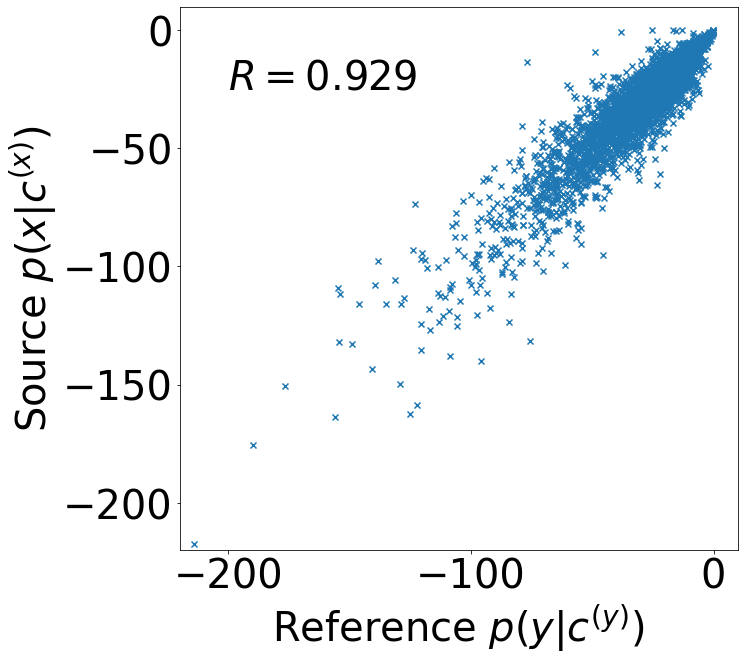}
        \caption{Cond.~prob.}
        \label{fig:pyc_corr_true}
    \end{subfigure}
    \begin{subfigure}{0.24\textwidth}
        \includegraphics[width=\linewidth]{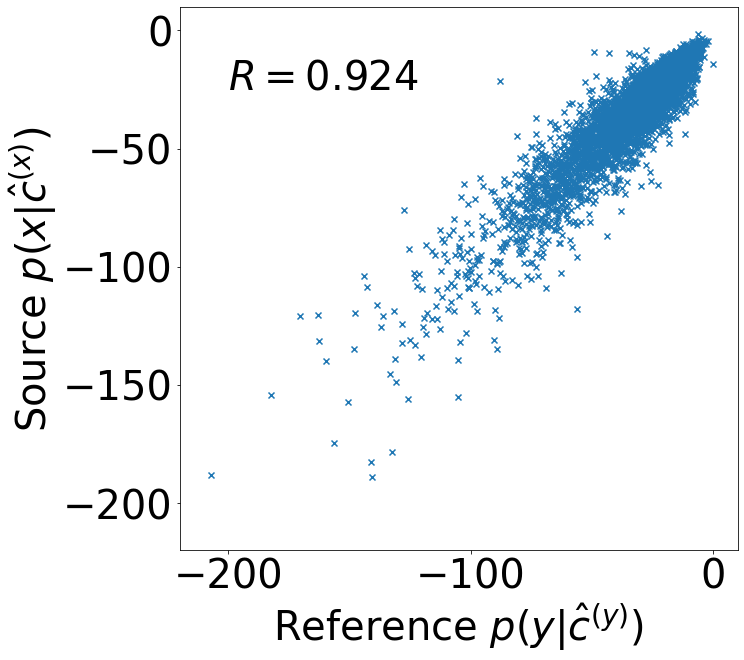}
%        \caption{Conditional probability with random context}
        \caption{Cond. prob.~(rand.~context)}
        \label{fig:pyc_corr_false}
    \end{subfigure}
    \caption{Source-target correlation of contextual \textsc{\textsc{pmi}} (a, b) and conditional probability (c, d), calculated based on the correct context (a, c) and wrong context that is randomly chosen from the dataset (b, d).
    The dataset is a subset of the training data from the English-Russian parallel corpus.
    Plots are for $4166$ sentence pairs in the dataset.
    }
    \label{fig:pmi_pyc_corr}
\end{figure*}

% We first confirm our methods do not sacrifice the translation performance measured with \textsc{bleu}. We then compare our methods with other baselines using a contrastive tests designed for evaluating context-aware translation~\cite{voita-etal-2019-good}.

\subsection{General translation performance measured by \textsc{bleu} scores}\label{sec:overall_bleu}
Table~\ref{tab:main_bleu} lists the performance of the models in terms of \textsc{bleu} scores. 
% \footnote{The hyper-parameter values for $T$ in reranking in \textsc{c-aware} beam, and $\beta$ in conditional/unconditional shallow fusion are $\{1000, 0.2, 250, 0.3, 0.1, 0.01\}$.}
% Hyper-parameter values are listed in Table~\ref{tab:hyper_params}.
Bayes DocReranker and our \textsc{c-aware} Rerank consistently outperformed the baseline SentTransformer, even when it used data augmentation by back-translation, while the other methods are just comparable to the baseline.
Althogh Bayes DocReranker performed the best among all the models, the comparison to Bayes DocReranker without context information (using $p_{\textsc{s-lm}}(\bm{y}_i)$ instead of $p_{\textsc{d-lm}}(\bm{y}_i|\bm{y}_{<i})$) reveals that most of the improvement is not obtained by the use of contexts. 
Back-translation did not contribute to \textsc{bleu} possibly because the original parallel data is already large and there was little room for improvement with additional pseudo data.
\begin{comment}
Without smoothing, our reranking with context-aware scores is comparable to baseline 1-to-1 and 2-to-2 decoders, while \textsc{c-aware} beam degrades \textsc{bleu} scores.
We later analyse the reason in detail in \S~\ref{sec:pmi_corr_models}.
%Both add-$\alpha$ and $T$-scaling smoothing are effective for recovering the performance and further improve \textsc{bleu} by 1.02 over 1-to-1.
\textsc{c-aware} beam and reranking with 
smoothing
% $\alpha$ and $T$ 
also outperform shallow fusion in the majority of cases.
%Observing conditional shallow fusion outperforming unconditional shallow fusion in all the configurations, we can assume conditional shallow fusion is able to capture contextual information to some extent.
\end{comment}

%Figure~\ref{fig:bleu_contrastive_hyperparams} depicts the trade-off of \textsc{bleu} scores and average scores of contrastive tests for En$\rightarrow$Fr translation with \textsc{c-aware} beam. This is a natural consequence of \textsc{c-score} that captures a valid translation in the two independent perspectives. We will design context-aware \textsc{blue} in the future.

\subsection{Results on contrastive test sets} \label{sec:contrastive_test}
Tables \ref{tab:disco} lists evaluation results (accuracy) of the contrastive tests with models using 30M monolingual data. 
% ToDo: 検討 D-LM のスコアを太字にする必要ある？
The highest scores on each column are in bold, and additionally, the higher one of the two \textsc{d-lm}-based scores is shown in bold.
The contrastive test include four test sets: \textit{deixis} is for person deixis, \textit{lex.c} is for lexical cohesion, \textit{ell.infl} is for inflection of Russian nouns caused by ellipsis in the source sentence, and \textit{ell.vp} is for verb ellipsis in English text which is not allowed in Russian.
Although the contrastive test is targeted at context-aware \textsc{nmt} models, it is possible to answer the contrastive questions by $\arg\max_y \textsc{pmi}(\bm{c}^{(\bm{y})},y)$ or $\arg\max_y p(\bm{y}|\bm{c}^{(\bm{y})})$.
Scores obtained by these two objectives are also reported in the table in addition to the scores obtained by SentTransformer.

% The proposed 
Our \textsc{c-score} outperforms all the context-aware models other than DocRepair. The performance of \textsc{c-score} is slightly worse than DocRepair for \textit{deixis} (2.2 points) and \textit{ell.infl} (4.0 points), while achieving large improvements for \textit{lex.c} (19.1 points) and \textit{ell.vp} (9.8 points) over DocRepair.

\textit{\textsc{d-lm} only} objectives achieve higher scores than \textsc{c-score}, except for \textit{ell.infl}.
% ToDo: 確認。流れがわかりづらいので修正↓
% although the \textsc{c-score} has the advantage of the information about $\bm{x}$ given.
This is not surprising because the choices in the tests are guaranteed to be valid as translation for the source sentences if given some appropriate context, so the questions can be solved without translation.
This result still indicates that the \textsc{d-lm} scores give good hints for tackling contextual ambiguities.
%This may be due to the difference of the distribution of the sentences in the monolingual corpora and the target-side of the parallel corpora.
The advantage of \textsc{c-score} over the SentTransformer is demonstrated by the excellent performance of \textsc{d-lm} in capturing contexts in translation.

% \textsc{c-score} with $T$-scaling shows comparable performance with 2-to-2 as a whole.
%\textsc{c-score} with add-$\alpha$ and $T$-scaling largely outperform 2-to-2 on the En-Ru lex.c test set, which may be because the test requires knowledge about low frequency words which language models can 

%\begin{figure*}
%    \centering
%    \includegraphics[width=14.cm]{fig/var_bleu_disco.pdf}
%    \caption{\textsc{bleu} and the average score of En$\rightarrow$Fr  contrastive test while varying hyperparameter values.}
%    \label{fig:bleu_contrastive_hyperparams}
%\end{figure*}

\begin{comment}
\begin{table}[t]
    \small
    \centering
    \begin{tabular}{lccc} \toprule
         &  En\rightarrow$Fr$ & En$\rightarrow$Ru & En$\rightarrow$Ja \\ \midrule
         Reranking \\
         \hfill $\alpha$ & 100 & 1000 & 100 \\
         \hfill $T$& 0.1 & 0.2 & 0.4 \\
         \textsc{c-aware} beam \\
         \hfill $\alpha$ & 150 & 250 & 100 \\
         \hfill $T$& 0.1 & 0.3 & 0.4 \\
         Shallow Fusion ($\beta$) \\
         \hfill Conditional & 0.01 & 0.1 & 0.1 \\
         \hfill Unconditional & 0.01 & 0.01 & 0.06 \\ \bottomrule
    \end{tabular}
    \caption{Hyper-parameters: $\alpha$ and $T$ for reranking and \textsc{c-aware} beam and $\beta$ for conditional/unconditional shallow fusion.}
    \label{tab:hyper_params}
\end{table}
\end{comment}

\subsection{On translation efficiency}

The inference speed depends mainly on the model size and beam size.
In our experiments on a single TITAN Xp GPU, SentTransformer decoded the fastest at 66 sents/sec, followed by DocTransformer that ran in 40 sents/sec.
DocRepair ran in about 28 sents/sec, slightly slower because it decodes in two passes.
\textsc{c-aware} Rerank and Bayes DocReranker were about 4.3 sents/sec and 7.7 sents/sec respectively.
We expect that these models would be accelerated by using a language model with a better cache mechanism (e.g. TransformerXL~\cite{dai-etal-2019-transformer}).
\textsc{c-aware} Beam ran in about 13 sents/sec.\footnote{Note that the running time of \textsc{nmt} decoding also depends on the degree of parallelism, and for \textsc{c-aware} Beam, decoding multiple sentences in parallel is less trivial since it demands that all the previous sentences in the document are translated by the time it starts to translate the current one. In our experiments, assuming a practical scenario where a large number of users input their documents for translation, we translate multiple documents in parallel so that multiple sentences from different documents can be translated in parallel.}
We leave thorough analysis on speed/performance trade-offs to future work.

\subsection{\textsc{pmi} correlation analysis}

%\begin{comment}
\begin{figure*}
    \centering
    \includegraphics[width=11cm]{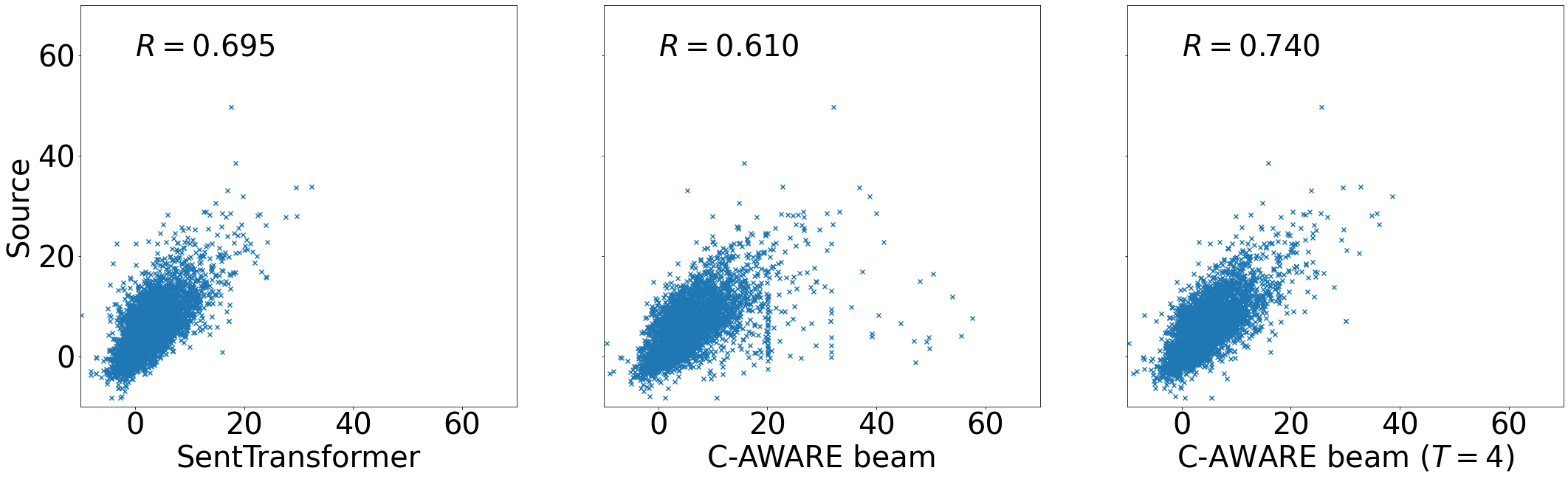}
    \caption{Correlation of contextual \textsc{pmi} between the source sentences (from the training data) and the outputs of some models (SentTransformer, \textsc{c-aware} beam without $T$-scaling, and \textsc{c-aware} beam with $T$-scaling of $T=4$).}
    \label{fig:pmi_corr_models}
\end{figure*}
%\end{comment}

In \S~\ref{sec:contrastive_test} we have confirmed the effectiveness of \textsc{pmi} as a measure of a valid translation given context using contrastive tests.
To gain a deeper insight into how well \textsc{pmi} conveys semantic connections between the current sentence and its context, we analyze the correlation of \textsc{pmi} between source and target sentences.
% and provide some analysis and visualizations.

\subsubsection*{\textsc{pmi} correlation between source and target}
The main result we show in this section is that the \textsc{pmi} of the source and target correlate well.
This is important because this supports the idea that \textsc{pmi} is a language-independent measure of the connection between the current sentence and its context.

Although we have discussed only target-side $\textsc{pmi}(\bm{c}^{(\bm{y})}, \bm{y})$ defined by Eq.~\ref{eq:c_pmi}, we can compute the source-side $\textsc{pmi}(\bm{c}^{(\bm{x})},\bm{x})$ in the same way.
Given a document-level parallel corpus, we measure a correlation between $\textsc{pmi}(\bm{c}^{(\bm{x})},\bm{x})$ and $\textsc{pmi}(\bm{c}^{(\bm{y})}, \bm{y})$ for  each sentence pair $(\bm{x},\bm{y})$  in the corpus.
% for every sentence pair $(\bm{x},\bm{y})$ in the corpus, a tuple $(\textsc{pmi}(\bm{c}^{(\bm{x})},\bm{x}), \textsc{pmi}(\bm{c}^{(\bm{y})}, \bm{y}))$.
% These tuples give us the correlation of \textsc{pmi} between source and target.

Figure~\ref{fig:corr_pmi} shows the \textsc{pmi} correlation for about $4000$ 
% En-Ru 
sentence pairs taken from the dev data.
The pairs of \textsc{pmi} values are computed using English and Russian language models trained on the training data.
%Each point of the scatter plot corresponds to a sentence pair.
%The y-coordinate is the contextual \textsc{pmi} of the source sentence and the x-coordinate is that of the target sentence.
We observe a clear correlation between source and target, which agrees with the intuition that if the target sentence matches well in the context, so does the source sentence.
What is also obvious in Figure~\ref{fig:corr_pmi} is that most of the points lay in the first quadrant where both the source and target contextual \textsc{pmi} is greater than $0$, which is explained by the simple intuition that most sentences should have positive co-occurrence relation with their contexts.
This behavior is lost when computing the contextual \textsc{pmi} using an incorrect context $\tilde{c}$ randomly chosen in the dataset as shown in Figure~\ref{fig:corr_pmi_false}.

The effectiveness of \textsc{pmi} as a measure of the valid translation of the current sentence given context is further emphasized when compared to the conditional probability $p(\bm{y}|\bm{c}^{(\bm{y})})$, which could be an alternative measure of how suitable $\bm{y}$ is in the context as described in \S~\ref{sec:shallow_fusion}.
%Here we show that \textsc{pmi} brings more contextual information than the conditional probability does by analysing the conditional probability correlation between source and target.
Figure~\ref{fig:pyc_corr_true} and \ref{fig:pyc_corr_false} are the conditional probability version of Figure~\ref{fig:corr_pmi} and \ref{fig:corr_pmi_false}: $(p(\bm{x}|\bm{c}^{(\bm{x})}),p(\bm{y}|\bm{c}^{(\bm{y})}))$ for each sentence pair $(x,y)$ in the same dataset are plotted in Figure~\ref{fig:pyc_corr_true} and the same tuples but with random contexts are plotted in Figure~\ref{fig:pyc_corr_false}.
%Each point on Figure~\ref{fig:pyc_corr_true} corresponds to a sentence pair $(x,y)$ and the y-coordinate of the point is the source-side conditional probability $\log p(x|c^{(x)})$ whereas the x-coordinate is the target-side conditional probability $\log p(\bm{y}|\bm{c}^{(\bm{y})})$.
%We compute the coordinates of the points on Figure~\ref{fig:pyc_corr_false} $p(x|c^{(x)})$ and $p(\bm{y}|\bm{c}^{(\bm{y})})$ using wrong context $\hat{c}^{(x)}$ and $\hat{c}^{(y)}$ randomly chosen from the dataset.
% As shown in the figures, 
Unlike the contextual \textsc{pmi} correlation, conditional probability correlation remains high even when we give wrong contexts.
This is because the conditional probability of a sentence is highly affected by how frequently 
the sentence is observed regardless of context; 
% for example, 
if the source sentence is written with common expressions, then so is the target sentence and they are likely to be observed regardless of the context.

%\begin{comment}
\subsubsection*{Analysis of the model outputs} \label{sec:pmi_corr_models}
\textsc{pmi} correlation gives us a good explanation of how \textsc{c-aware} beam without $T$-scaling fails.
% (see Table~\ref{tab:main_bleu}).
We plot the \textsc{pmi} correlation between the source sentences and their translations obtained with \textsc{nmt} models (Figure~\ref{fig:pmi_corr_models}).
We can find some outliers in the bottom right area of the plot for \textsc{c-aware} beam without $T$-scaling, which is 
% suspected to be
the cause of the low correlation coefficient $R=0.610<R_{\mathrm{src-ref}}=0.695$.
This result suggests that \textsc{c-aware} beam without $T$-scaling chooses some tokens based on excessively high token-wise \textsc{pmi}, which breaks some translations resulting in the low \textsc{bleu}.
Translation of the SentTransformer shows a higher correlation with the source texts than the reference translation (Figure~\ref{fig:corr_pmi}).
One possible explanation for this is alignment errors in the corpus: although worse than the reference translations in quality, outputs of SentTransformer are considered to be perfectly aligned to the source sentences.
\textsc{c-aware} beam with $T$-scaling ($T=4$) seems to solve this issue and achieves the highest \textsc{pmi} correlation $R=0.740$.
%\end{comment}

\section{Related Work} \label{sec:related_work}
The effectiveness of incorporating context into translation was shown in earlier literature on document-level \textsc{nmt}~\cite{tiedemann-scherrer-2017-neural,bawden-etal-2018-evaluating} using the single encoder architecture.
Multi-encoder architectures were explored to better capture contextual information~\cite{wang-etal-2017-exploiting-cross,tu-etal-2018-learning,jean2017does,miculicich-etal-2018-document,voita-etal-2018-context,bawden-etal-2018-evaluating,maruf-haffari-2018-document,maruf-etal-2019-selective,kang-etal-2020-dynamic,zhang-etal-2020-long}.
% As the multi-encoder models often show a comparable or better performance over the simple yet effective single-encoder baseline~\cite{tiedemann-scherrer-2017-neural}, 
However, since parallel data is often constructed by picking up reliable sentential alignments from comparable documents, document-level sentence-aligned parallel data for training these document-level \textsc{nmt} models are expensive to obtain and available in only a few domains and language pairs~\cite{sugiyama-yoshinaga-2019-data}.

% Since these approaches rely on expensive document-level parallel data, recent studies have started to focus on modeling contexts using document-level monolingual data.

Recent studies have therefore started to focus on modeling contexts using document-level monolingual data. The current approaches are grouped into three categories: data augmentation via back-translation~\cite{sugiyama-yoshinaga-2019-data}, a post-editing model~\cite{voita-etal-2019-context}, and modeling document-level fluency via document-level \textsc{lm}s~\cite{stahlberg-etal-2019-cued,yu-etal-2020-better,jean-cho-2020-log}. In what follows, we review these approaches in detail.

\citet{sugiyama-yoshinaga-2019-data} reported that the data augmentation by back-translation~\cite{sennrich-etal-2016-improving} enhances a document-level \textsc{nmt} model with a single encoder architecture in low-resource settings. However, we have obtained limited improvements in our settings (Table~\ref{tab:main_bleu} and Table~\ref{tab:disco}). Moreover, this approach is expensive since it learns a document-level \textsc{nmt} model from a massive amount of pseudo parallel data.

\citet{voita-etal-2019-context} proposed DocRepair, a context-aware post-editing model that corrects outputs of a sentence-level \textsc{nmt} model. Because DocRepair ignores the confidence of the first-stage sentence-level translation and possible alternative translations, it can miscorrect outputs of the sentence-level \textsc{nmt} model when they are irregular but correct. Moreover, when we change the target sentence-level \textsc{nmt} model, the accompanying post-editing model must be trained from its outputs.
% DocRepair only looks at the outputs of the first stage sentence-level \textsc{nmt} and thus cannot access fine-grained information behind the surface such as the model's confidence and possible alternative translations.
% This limitation potentially causes miscorrection when the output of the sentence-level \textsc{nmt} is irregular but correct. 
Our approaches, on the other hand, attempt a more ``soft'' revision, taking into account the output probabilities, i.e., confidence of the sentence-level \textsc{nmt}, and can perform context-aware decoding with any sentence-level \textsc{nmt} model, reusing a pre-trained document-level \textsc{lm}.

\citet{stahlberg-etal-2019-cued} and \citet{yu-etal-2020-better} utilize a document-level \textsc{lm} to model document-level fluency of outputs; these approaches are similar to shallow fusion~\cite{gulcehre2015using}\footnote{Our work is also related to shallow fusion~\cite{gulcehre2015using}, in which token-wise probabilities output by an \textsc{nmt} model and a sentence-level \textsc{lm} are combined to be used as translation scores in decoding.
The theoretical background of shallow fusion and our \textsc{c-score} are different: in shallow fusion, the \textsc{lm} is intended to promote fluency of translations, whereas in our \textsc{c-score}, we use the probability ratio of two \textsc{lm} probabilities which only provides contextual difference and fluency is still left to the translation model.} with document-level \textsc{lm} (\S~\ref{sec:shallow_fusion}), although they perform a document-level reranking of translation hypotheses generated for individual source sentences by using sentence-level \textsc{nmt}. In particular, Yu's formulation has a probabilistic foundation like our approaches, and additionally utilizes a backward translation model.
% ここまで細かく説明しなくても切れそう
% they repeatedly substitute the top-1 hypotheses by lower ranked hypotheses to gradually improve document-level \textsc{lm} score. 
% \citet{yu-etal-2020-better} proposed another document-level reranking using a reranker consisting of a document-level \textsc{lm}, a forward translation model and a backward translation model.
Although their formulation brings a significant improvement in \textsc{bleu} (Table~\ref{tab:main_bleu}), the score is not obtained by better document-level translation; the comparable \textsc{bleu} score of the no-context version of the method (Table~\ref{tab:main_bleu}) and the results of the contrastive tests (Table~\ref{tab:disco}) reveal that the improvement is mostly due to the context-agnostic language model prior and the backward translation model. As we have discussed in \S~\ref{sec:shallow_fusion}, document-level \textsc{lm} scores prefer tokens which frequently appear regardless of context and are unlikely to lead to better document-level translation. Moreover, their method requires training a back-translation model corresponding to 
the target sentence-level \textsc{nmt} model.

% As we have discussed in \S~\ref{sec:shallow_fusion} and the experimental results and analysis, \textsc{c-score} captures contextual information better than the shallow fusion score with extended context.

 Finally, we noticed that \citet{jean-cho-2020-log} (which appeared after the preprint version of this paper~\cite{sugiyama2020contextaware}\footnote{This preprint is submitted to and rejected from EMNLP 2020; the interested reader may refer to this paper for experiments on other language pairs such as English to French and English to Japanese translation.} had been submitted) have reached a formulation that is very similar to the one presented in this paper by reformulating a noisy channel model of Bayes DocReranker~\cite{yu-etal-2020-better}.
Concrete differences between our work and theirs include the fact that we conducted thorough analysis on the performance of different decoding strategies (not only beam search but also reranking). We also interpreted the subtraction of \textsc{lm} scores as point-wise mutual information and analyzed it by observing \textsc{pmi} correlation between source and target \textsc{pmi} to deepen the understanding of the formulation.

% ↓ 実用的なMTにはリランカが含まれることが普通で、リランカの構成要素として逆翻訳モデルなどの補助ネットワークを学習するのは一般的なので、わざわざ指摘することではない？
% In addition, the model requires to train multiple auxiliary models including a back-translation model, which incurs additional training cost compared to our method.

\section{Conclusions}
We present an approach to context-aware \textsc{nmt} based on \textsc{pmi} between the context and the current sentence.
We first provide the formulation of the objective, \textsc{c-score}, and the computation process of the \textsc{c-score} using a sentence-level translation model and a document-level language model.
We investigate two search methods, reranking and beam search, and evaluate the methods for English-Russian translation.
We also provide some analysis and visualization to better understand the nature of \textsc{pmi} between the context and the current sentence.

We plan to design context-aware \textsc{bleu} using \textsc{pmi} for evaluating context-aware \textsc{nmt} models. We will evaluate our method on non-autoregressive \textsc{nmt}~\cite{gu2014nonauto}. We will release all code and data to promote the reproducibility of  results.\footnote{\url{http://www.tkl.iis.u-tokyo.ac.jp/~sugi/NAACL2021/}}

\section*{Acknowledgements}
We thank anonymous reviewers for their valuable comments. We also thank Joshua Tanner for proofreading this paper. We also thank Masato Neishi for technical advice on implementations of neural machine translation. The research was supported by NII CRIS collaborative research program operated by NII CRIS and LINE Corporation.

\bibliography{naacl2021}
\bibliographystyle{acl_natbib}

\end{document}